\documentclass[10pt,twocolumn,letterpaper]{article}
\pdfoutput=1

\usepackage{iccv}
\usepackage{times}
\usepackage{graphicx}
\usepackage{tabularx}
\usepackage{amsmath}
\usepackage{amssymb}
\usepackage{bm}
\usepackage{multirow}
\usepackage{soul}
\usepackage{makecell}
\usepackage{pifont}
\usepackage{booktabs}
\usepackage{caption}
\usepackage{subcaption}
\usepackage{authblk}

\newcommand{\mv}[1]{\textcolor{blue}{mv:\xspace#1}}
\newcommand{\steven}[1]{\textcolor{cyan}{steven:\xspace#1}}

\newcommand{\boldstart}[1]{\noindent\textbf{#1}}
\newcommand{\suppress}[1]{}

\newcommand{\methodname}{\text{ODAM}\xspace}
\newcommand{\se}{\mathrm{SE(3)}\xspace}

\DeclareMathOperator*{\argmax}{arg\,max}

\usepackage[pagebackref=true,breaklinks=true,letterpaper=true,colorlinks,bookmarks=false]{hyperref}

\iccvfinalcopy 

\def\iccvPaperID{2885} 

\ificcvfinal\fi

\begin{document}

\title{ODAM: Object Detection, Association, and Mapping using Posed RGB Video}



\makeatletter
\renewcommand\AB@affilsepx{, \protect\Affilfont}
\makeatother

\author[1, 2]{Kejie Li}
\author[2]{Daniel DeTone}
\author[2]{Steven Chen}
\author[2]{Minh Vo}
\author[1]{Ian Reid}
\author[3]{Hamid Rezatofighi}
\author[2]{Chris Sweeney}
\author[2]{Julian Straub}
\author[2]{Richard Newcombe}

\affil[1]{The University of Adelaide}
\affil[2]{Facebook Reality Labs Research}
\affil[3]{Monash University}
\maketitle
\ificcvfinal\thispagestyle{empty}\fi

\begin{abstract}

Localizing objects and estimating their extent in 3D is an important step towards high-level 3D scene understanding, which has many applications in Augmented Reality and Robotics. 
We present \methodname, a system for 3D \textbf{O}bject \textbf{D}etection, \textbf{A}ssociation, and \textbf{M}apping using posed RGB videos. 
The proposed system relies on a deep learning front-end to detect 3D objects from a given RGB frame and associate them to a global object-based map using a graph neural network (GNN). 
Based on these frame-to-model associations, our back-end optimizes object bounding volumes, represented as super-quadrics, under multi-view geometry constraints and the object scale prior. 
We validate the proposed system on ScanNet where we show a significant improvement over existing RGB-only methods.
\end{abstract}


\section{Introduction}


Endowing machine perception with the capability of inferring 3D object-based maps brings AI systems one step closer to semantic understanding of the world. 
This task requires building a consistent 3D object-based map of a scene. We focus on the space between the category-level semantic reconstructions \cite{mccormac2017semanticfusion} and object-based maps with renderable dense object models \cite{maninis20vid2cad, runz2020frodo} and represent objects by the 3D bounding volumes from posed RGB frames.
As an analogy to the use of 2D bounding boxes (BBs) in images, a 3D bounding volume presents a valuable abstraction of location and space, enabling for example, 
object-level planning for robots~\cite{ekvall2008robot, harada2014validating}, learning scene-level priors over objects~\cite{wang2019planit}, or anchoring information on object instances.
A robust way of inferring bounding volumes and associated views of individual objects in a scene is a stepping stone toward reconstructing, embedding and describing the objects with advanced state-of-the-art methods such as NeRF~\cite{mildenhall2020nerf}, and GRAF~\cite{schwarz2020graf}, which commonly assume a set of associated frames observing an object or part of a scene that can be obtained from the proposed reconstruction system.

\begin{figure}[t]
    \centering
    \includegraphics[width=\linewidth]{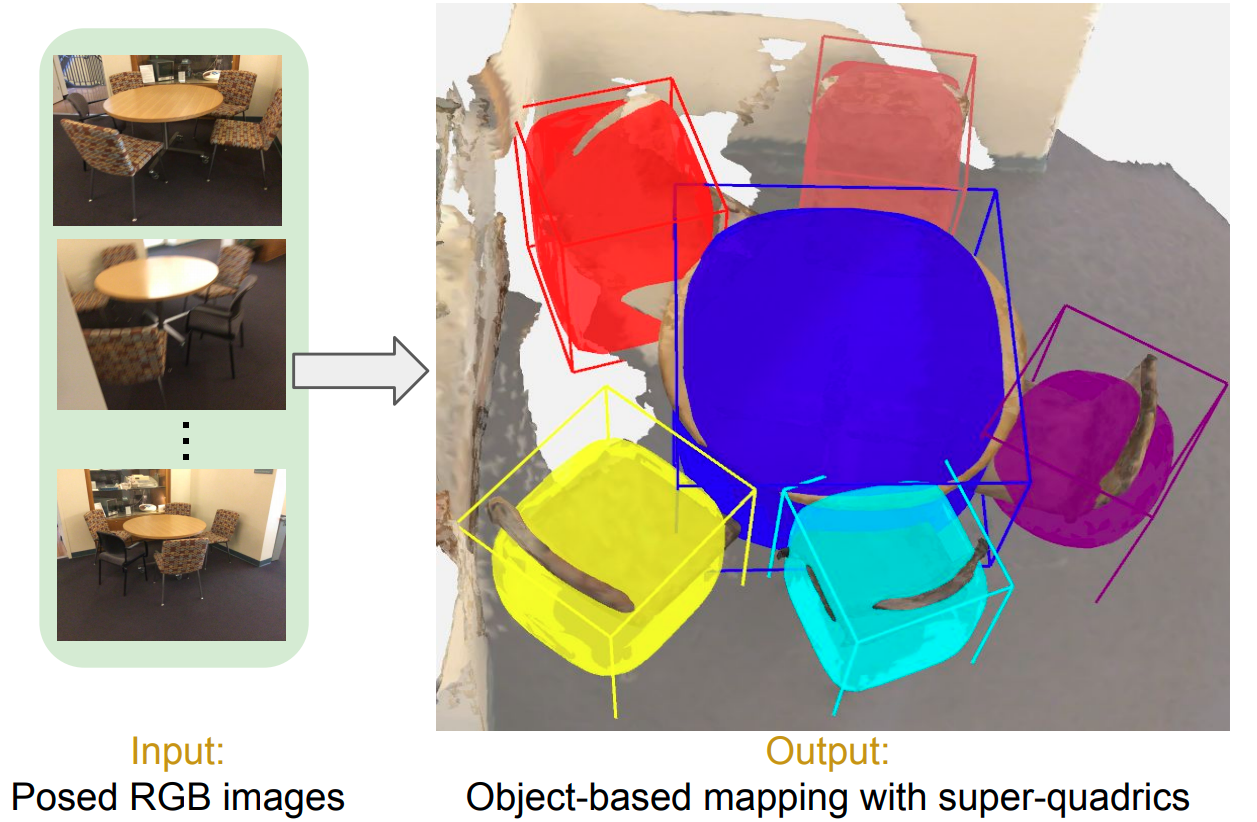}
    \caption{\methodname overview. Given a posed RGB video, \methodname estimates oriented 3D bounding volumes of objects represented by super-quadrics in a scene.}
    \label{fig:teaser}
\end{figure}

Nevertheless, this task of localizing objects and estimating their extents in 3D using RGB-only videos presents a number of challenges. First, despite the impressive success of deep learning methods for 2D object detectors~\cite{carion2020detr,he2017mask,redmon2016you}, recent efforts that formulate 3D object mapping as a single-view 3D detection problem~\cite{brazil2020kinematic,li2020mo,mousavian20173d} suffer from accuracy due to the depth-scale ambiguity in the perspective projection (as demonstrated empirically in Sec.~\ref{subsec:exp_ablation}).
Second, unlike estimation of 3D points from multiple 2D observations that has been studied extensively in SfM and SLAM~\cite{cadena2016past,engel2017direct, klein2007parallel, mur2015orb,tomasi1992shape}, there has been little work and consensus on how to leverage multi-view constraints for 3D bounding volume location and extent~\cite{nicholson2018quadricslam, yang2019cubeslam}.
Specifically, the representation for 3D volume and how to formulate a suitable energy function remain open questions. 
Third, the crucial problem that needs to be solved prior to multi-view optimization is the associations of detections of individual 3D object instances from different viewpoints, where unlike SfM or SLAM incorrect association noticeably biases the 3D object localization. However, this problem is under-explored for cluttered indoor environments, where specific problems such as having multiple objects with near identical visual appearance and heavy occlusion (\eg, multiple chairs closely arranged in a room as can be seen in Fig.~\ref{fig:scan2cad_result}) are commonplace. Depth ambiguity and partial observations complicate the data association problem.

\begin{figure*}[t]
    \centering
    \includegraphics[width=\textwidth]{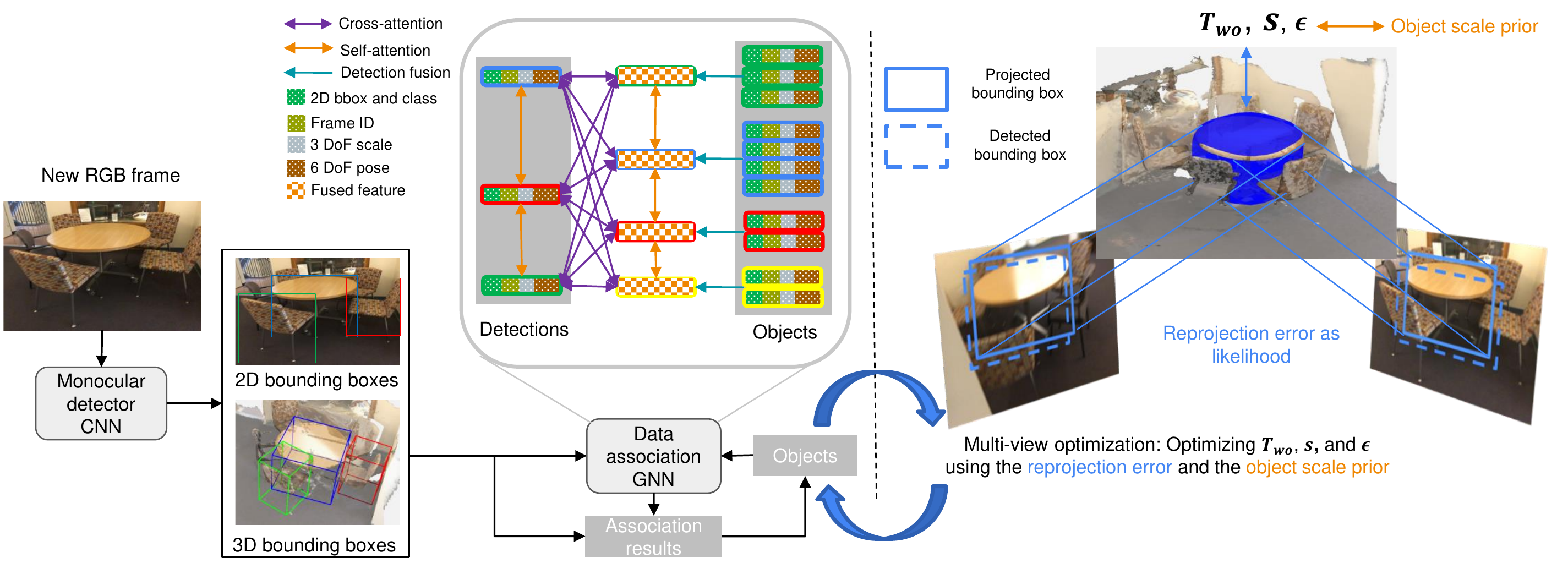}
    \caption{\methodname pipeline. Given a new RGB frame, a single-view detector  (Sec.~\ref{subsec:detector}) detects objects at the current frame. A GNN takes as inputs the new detections and existing objects in the map to predict the assignment matrix (Sec.~\ref{subsec:association}). 
    Concurrent with the front-end of the system (\ie detection and association), the location and extent of each object are represented by a super-quadric, which is optimized using the associated 2D BBs and category-conditioned scale prior (Sec.~\ref{subsec:optimization}).
    }
    \label{fig:system_pipeline}
\end{figure*}

We propose \methodname, a novel framework that incorporates a deep learning front-end and multi-view optimization back-end to address 3D object mapping from posed RGB videos.
The advantage of using RGB-only over RGB-D is significantly less power consumption. 
We assume the poses of the images are known; these are readily available with modern mobile/AR devices.
The front-end first detects objects of interest and predicts each object's 2D attributes (2D BB, object class), as well as its 3D BB parameterized by 6~Degree-of-Freedom (DoF) rigid pose and 3~DoF scale given a single RGB frame as shown in Fig.~\ref{fig:system_pipeline}.
The primary use of the 3D attributes for each detection is to facilitate data association between a new frame and the current global 3D map. 
Concretely we develop a graph neural network (GNN) which takes as inputs the 2D and 3D attributes of the current frames detections and matches them to existing object instances in the map. 
The front-end of can run 6~fps on average on a modern GPU on cluttered scenes such as those in ScanNet~\cite{dai2017scannet}.

The back-end of \methodname is a multi-view optimization that optimizes each object's oriented bounding volume represented by a super-quadric surface given multiple associated 2D bounding box (BB) observations.
Previous object-level SLAM frameworks have adopted either cuboids \cite{yang2019cubeslam} or ellipsoids~\cite{hosseinzadeh2019real, nicholson2018quadricslam} as their object representation, but they are often not a good model for the extent of a generic object as depicted in Fig.~\ref{fig:representation_comparison}.
Super-quadric -- a unified representation for shape primitives including cuboids, ellipsoids, and cylinders -- permits blending between cuboids and ellipsoids (and cylinders) and can therefore provide a tight bounding volume for the multi-view optimization. 
While super-quadric has been used to fit point cloud data~\cite{pentland1986parts, pentland1991closed, solina1990recovery} or recently parse object shapes from a single image using a deep network~\cite{Paschalidou2019CVPR}, we present the first approach to optimize super-quadrics given multiple 2D BB observations to the best of our knowledge.
Besides the representation, we realize that the 2D BBs given by the object detector are not error free due to occlusions in cluttered indoor environments.
We incorporate category-conditioned priors in the optimization objective to improve the robustness.

\boldstart{Contribution.} Our contributions are threefold: (1) we present \methodname, a novel online 3D object-based mapping system that integrates an deep-learning front-end running at 6~fps, and a geometry-based back-end. \methodname is the current best performing 3D detection and mapping RGB-only systems for complex indoor scenes in ScanNet~\cite{dai2017scannet}; 
(2) we present a novel method for associating single-view detections to the object-level. Our association employs a novel attention-based GNN taking as inputs the 2D and 3D attributes of the detections; (3) we identify the limitations of common 3D bounding volume representations used in multi-view optimization and introduce a super-quadric-based optimization under object-scale priors which shows clear improvements over previous methods.

\section{Related Work}
\noindent\textbf{3D object-based mapping.}
Approaches to 3D object-based mapping can be broadly classified into two categories: learning-based and geometry-based.
The first category mostly extends existing 2D detectors to also output 3D bounding box from single images~\cite{li2018,mahendran2018,massa2016PoseRegression,mousavian20173d,tulsiani2015PoseClassification,xiao2019}. 
If a video sequence is available, the single-view 3D estimations can be fused using a filter or a LSTM to create a consistent mapping of the scene~\cite{brazil2020kinematic,hu2019joint,li2020mo}. Yet, the fused 3D detections might not satisfy multi-view geometry constraints.
While the front-end of our proposed system is inspired by these learning-based approaches, we notice that single-view 3D inference is inaccurate because the inherent scale and depth ambiguity in 2D images and solve this issue with a back-end multi-view optimization.
The second category focuses on estimating the bounding volume of an 3D object given 2D detections from multiple views in a similar way to the reprojection error used in SfM and SLAM.
~\cite{crocco2016structure,rubino20173d} estimate the 3D ellipsoid representing the bounding volume of an object by minimizing the discrepancies between the projected ellipsoid and the detected 2D bounding boxes.
QuadricSLAM~\cite{nicholson2018quadricslam} represents objects as dual quadrics to be optimized with a novel geometric error and extends it to a full SLAM system.
CubeSLAM~\cite{yang2019cubeslam} uses 3D cuboids in the optimization and enforces reprojection error on the vertices of the 3D cuboids.
Our proposed multi-view optimization uses super-quadric -- a representation subsuming both ellipsoid and cuboid -- with the energy function formulation using joint 2D BBs and a scale prior constraint.

\noindent\textbf{Object-based mapping with 3D shape estimation.}
Extending beyond 3D oriented bounding boxes, several works focus on estimating dense object shapes via shape embedding~\cite{runz2020frodo} or CAD model retrieval~\cite{maninis20vid2cad} given posed RGB video. 
RfD-Net~\cite{nie2020rfd} explores completing full object shapes by first detecting 3D bounding boxes followed by a shape completion network for each detected object from 3D point cloud.
Although these methods estimate high-resolution object mapping, they require known 3D shape priors.
We do not assume prior knowledge of CAD models and instead focus on instance-agnostic pose estimation. 


\noindent\textbf{Associating detection across video frames.}
Associating object detections of the same 3D object instance across multiple frames has been studied in different contexts, and most prominently in the context of Multiple Object Tracking (MOT). 
MOT focuses on tracking dynamic objects (\eg cars and pedestrians) and often follows frame-to-frame paradigm by heavily exploiting the discriminative visual appearance features of objects~\cite{hu2019joint,tang2017multiple}. 
Until the recent end-to-end tracking approaches~\cite{braso2020learning, meinhardt2021trackformer,sun2020transtrack}, most approaches rely on simple motion continuity priors~\cite{dehghan2015gmmcp,milan2013continuous,wu2012coupling} to link instances. 
More closely related to \methodname is Weng~\etal~\cite{weng2020gnn3dmot} that uses a GNN to learn a matching cost given both point cloud and RGB images. 
Our proposed framework takes as input RGB-only images, and hence solves a more difficult but ubiquitous problem.
Moreover, rather than associating people or cars with discriminative visual appearance, we focus on indoor static object mapping where we associate a sets of highly repetitive objects from drastically different views (\eg front and back of a chair) in a frame-to-model fashion. 
Prior methods working for indoor environments resort to handcrafted matching by IoU~\cite{li2020mo} or SLAM feature point matching~\cite{hosseinzadeh2019real, yang2019cubeslam}, whereas we learn the matching using the GNN.



\noindent\textbf{3D detection from 2.5D and 3D input.}
There are several methods for 3D detection using 3D input~\cite{qi2020imvotenet,shi2019pointrcnn,yang2019std} or single RGBD images~\cite{chen2017multi,song2016deep}. 
Methods for instance segmentation~\cite{he2020dyco3d,hou2019sis,yang2019learning} given 3D input are also able to produce 3D bounding boxes. 
Since depth information directly resolves the scale ambiguity of an observed object, these methods solve a strictly easier problem.

\vspace{-0.25cm}
\section{Method}
The goal of \methodname is to localize objects and estimate their bounding volume accurately in 3D given posed \textit{RGB-only} image sequence. 
As shown in Fig.~\ref{fig:system_pipeline}, given an RGB frame, the front-end first detects objects and predicts their 2D and 3D attributes in the camera coordinate frame (Sec.~\ref{subsec:detector}).
These detections are associated to existing object instances in the map or become a new object instance by solving an assignment problem using a GNN (Sec.~\ref{subsec:association}).
Given the association from the front-end, our back-end system optimizes a super-quadric surface presentation of each object from multiple associated 2D BB detections and category-conditioned object-scale priors from all associated views. (Sec.~\ref{subsec:optimization}). 

\subsection{Single-view 2D and 3D Object Detection}\label{subsec:detector}
\methodname first detects objects of interest given a new RGB frame.
Our detector is a single-view 3D detector that estimates not only 2D attributes -- 2D BB and object class -- but also 3D attributes -- translation $\bm{t}_{co}$, rotation $\bm{R}_{co}$, and 3D BB dimensions $\bm{s}$ with respect to the local camera coordinate frame. 
Specifically, we estimate $\bm{t}_{co}$ by predicting its depth and 2D center on the image. $\bm{R}_{co}$ is formulated as a classification on three Euler angles. 

\subsection{Association of Detections to Object-based Map}\label{subsec:association}
Detections from the single-view detector are matched to existing object instances in the map using an attention-based GNN as opposed to handcrafted data association algorithms used in prior art~\cite{li2020mo,maninis20vid2cad}.
The benefit of using a GNN for data association is twofold. 
First, different attributes (\eg 2D BB, 3D BB, object class) can be taken as joint input to the network to extract more discriminative features for matching. 
Second, instead of only considering pair-wise relationships in handcrafted data association methods, the attention mechanism in GNN aggregates information from other nodes in the graph for more robust matching. 
Thus, our GNN can infer the association of an object detection from the full set of objects in the scene, as visualized in Fig.~\ref{fig:system_pipeline}.

Our association graph is implemented as a GNN where each node is a feature descriptor comprising of 2D and 3D information of an object detection and edges connect (1) among previously associated detections of an object in the object fusion; (2) a new detection to other detections  and a fused object feature vector to other object feature vectors for self-attention; (3) a new detection to fused object feature vectors for cross-attention, as shown in Fig.~\ref{fig:system_pipeline}.
This graph predicts a matching between a set of input detection and existing objects in the map.  
For every object in the map, we fuse its descriptors from all associated views using self-attention GNN layers. These fused descriptors are matched to the descriptor of the newly detected objects using self- and cross-attention GNN layers.

\noindent\textbf{Input detection features.}
The $m^{th}$ new detection at frame $t$ is represented by a feature descriptor $\bm{d}^t_m \in \mathbb{R}^{16}$ comprising the frame ID, the detected 2D BB, object class, detection score, 6~DoF object pose, and 3~DoF object scale given by the monocular detector.
The $n^{th}$ object instance is represented by a set of associated detections in previous frames $\{ \bm{d}^{t_0}_n, \bm{d}^{t_1}_n, ..., \bm{d}^{t_l}_n\}$, where $\bm{d}^{t_l}_n$ is a previously associated detection of the $n^{th}$ object instance at frame $t_l$. 
To facilitate association of the detections in new RGB frame to the mapped objects, the detected 2D BB and 6~DoF object pose in $\bm{d}^{t_l}_n$ are replaced by the projection from the estimated 3D bounding volume to the current frame coordinate.

\noindent\textbf{Object fusion.}  
We first fuse all associated detections of a mapped object using a self-attention GNN into a single feature descriptor vector: 
\begin{align}
\bm{o}_n = f_d(\{ \bm{d}^{t_0}_n, \bm{d}^{t_1}_n, ..., \bm{d}^{t_l}_n\})\,,    
\end{align}
where $f_d(\cdot)$ is the self-attention GNN taking as input a set of previously associated detections of an object instance, and $\bm{o}_n \in \mathbb{R}^{256}$ is the fused feature vector for the object instance for data association. 
This step allows information across observations of the same object from different viewpoints to be aggregated before matching to the new detection at the current frame.

\noindent\textbf{Frame-to-model association.} 
After the fusion step, the frame-to-model data association becomes a bipartite matching problem, where the two disjoint subsets are the fused vectors of $m$ existing objects and $n$ new detections at the current frame $t$ respectively. 
This matching problem is solved by the second part of the GNN, which contains a stack of alternating self-attention layers aggregating information within the subset and cross-attention layers aggregating information from the other subset. 
The assignment matrix $\bm{M} \in \mathbb{R}^{m\times n}$ is computed as:
\begin{align}
\bm{M} = f_m(\{\bm{o}_0, \bm{o}_1, ..., \bm{o}_m\}, \{\bm{d}^t_0, \bm{d}^t_1, ..., \bm{d}^t_n\})\,,
\end{align} 
where $f_m(\cdot)$ is second part of the GNN taking as input objects' fused vectors $\bm{o}_n$ and new detections $\bm{d}^t$. 
Please refer to Sec.~\ref{subsec:details} and the supplementary material for more network and training details. 

\begin{figure*}
    \centering
    \begin{minipage}{0.6\textwidth}
    \centering
        \includegraphics[height=0.2\paperheight]{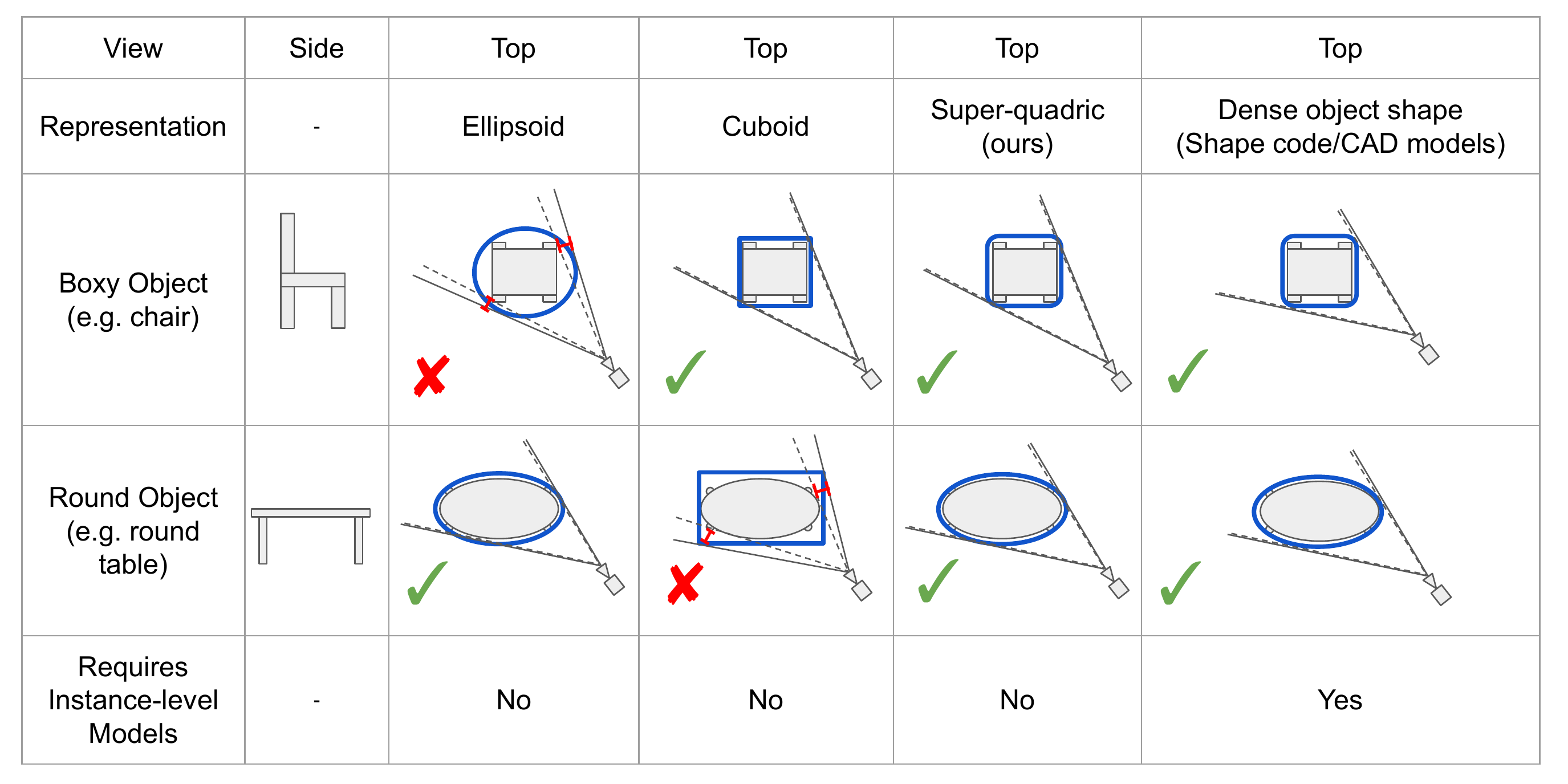}
        \caption{Limitations of object representations in multi-view optimization from 2D BB observations. Ellipsoid and cuboid are only suitable for a subset of objects. Dense representations require instance-level models. }
        \label{fig:representation_comparison}
    \end{minipage}
    \hfill
    \begin{minipage}{0.35\linewidth}
        \centering
        \includegraphics[height=0.2\paperheight,trim=10 10 0 0,clip]{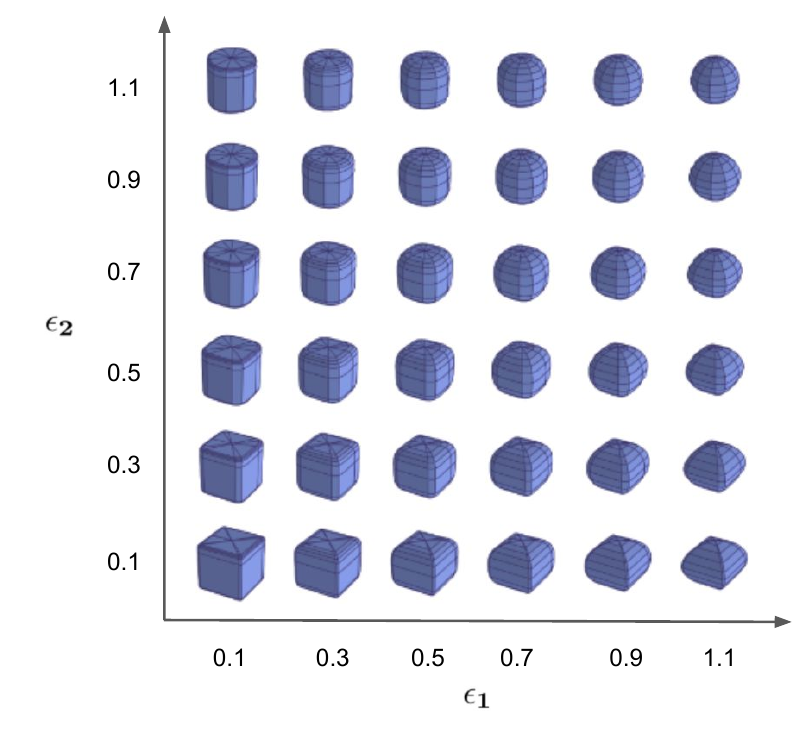}
        \caption{Super-quadric visualization. Different $\epsilon_{1,2}$ values include cuboids, ellipsoids, and cylinders (figure credit~\cite{superquadric_web}).}
        \label{fig:super_quadric}
    \end{minipage}
\end{figure*}

\subsection{Multi-view Optimization}\label{subsec:optimization}
Instead of relying on a single-view 3D detector to solve the ill-posed monocular 3D detection problem, we propose a multi-view optimization for accurate 3D object mapping given multiple associated 2D BBs obtained from the previous step (Sec.~\ref{subsec:association}). 
The key to the optimization is representing a bounding volume via a super-quadric with the realization that both ellipsoid and cuboid used in prior art are only suitable to a subset of object shapes. 
Specifically, given multiple 2D BBs, the estimated 3D bounding volume is a convex set bounded by the intersection region of all frustums.
As the number of viewpoints increases, the convex set converges to the convex hull (\ie the tightest convex set of the object shape). 
However, neither ellipsoid or cuboid is flexible enough to approximate the convex hull for generic objects. 
For instance, while an ellipsoid is suitable for round objects, it introduces inherent inconsistency when representing a box-like object as shown in Fig.~\ref{fig:representation_comparison}.
Super-quadric alleviates this issue by using the best fitted shape primitive in the family.
Although dense object shape representations (\eg shape codes~\cite{park2019deepsdf}, or CAD models~\cite{maninis20vid2cad} do not suffer from the inconsistency in projection, they require knowledge of instance-level object shape.

\noindent\textbf{Super-quadric formulation.}
We represent an object's bounding volume in 3D by a super-quadric. 
The canonical implicit function of a super-quadric surface has the following form~\cite{barr1981superquadrics}:
\begin{align}
    f(\bm{x}) = \left(\left(\frac{x}{\alpha_1}\right)^{\frac{2}{\epsilon_2}} + \left(\frac{y}{\alpha_2}\right)^{\frac{2}{\epsilon_2}}\right) ^ {\frac{\epsilon_2}{\epsilon_1}} + \left(\frac{z}{\alpha_3}\right) ^ {\frac{2}{\epsilon_1}}\,,
\end{align}
where $\bm{x}=[x,y,z]$ is a 3D point, $\bm{\alpha} = [\alpha_1$, $\alpha_2$, $\alpha_3]$ controls the scale on three axes (\ie the object's 3D dimensions), and $\epsilon_1$, and $\epsilon_2$ decide the global shape curvature. 
The shape transition from an ellipsoid to a cube controlled by $\epsilon_1$, and $\epsilon_2$ is visualized at Fig.~\ref{fig:super_quadric}.

A point $\bm{x}$ on the surface of a super-quadric can be transformed from the canonical coordinate to the world coordinate by a 6~DoF rigid body transformation matrix $\bm{T}_{wo} \in \se$.
Thus, a super-quadric in the world coordinate is parameterized by $\bm{\theta} \in \mathbb{R}^{11}$, comprising of $\bm{T}_{wo}$ (6~DoF to represent the rigid body transformation), and the 5 parameters of the super-quadric $\bm{\alpha}$ and $\epsilon_1$, $\epsilon_2$. 

\noindent\textbf{Optimization objective.}
The detected 2D BBs are inevitably inaccurate and noisy. 
While existing methods using multi-view constraints for 3D object mapping~\cite{hosseinzadeh2019real,maninis20vid2cad,nicholson2018quadricslam,yang2019cubeslam} only consider the reprojection error, we observe that prior knowledge on the object's 3D scale can improve the robustness of the estimation.
To incorporate the prior knowledge about object scale, we formulate
 object-based mapping as Maximum a Posterior (MAP) estimation of each object's super-quadric parameters $\bm{\theta}$. 
With the reprojection likelihood $P(\bm{b}_i | \bm{\theta})$ and the category-conditioned scale prior $P(\bm{\theta})$ the MAP problem is:
\begin{align}\label{eq:map}
    \argmax_{\bm{\theta}} P(\bm{\theta} | \bm{B}) &= \argmax_{\bm{\theta}} P(\bm{\theta}) \prod_i P(\bm{b}_i | \bm{\theta})\,,
\end{align} where 
$\bm{B} = \{\bm{b}_0, ..., \bm{b}_N\}$ is a set of $N$ associated 2D detected BBs. $\bm{b}_i$ is the detected 2D BB at frame $i$ described by its four corner points $[x_{\min}, x_{\max}, y_{\min}, y_{\max}]$.
Assuming zero-mean Gaussian noise on the 2D BB detection corners, the reprojection likelihood is 
\begin{align}
 P(\bm{b}_i|\bm{\theta}) &= \mathcal{N}(\bm{b}_i | \hat{\bm{b}_i}, \sigma^2 )\,,
\end{align}
where $\sigma$ is the assumed image noise and $\hat{\bm{b}_i}$ is the super-quadric's projection. It is computed as:
\begin{align}
\hat{\bm{b}_i} &= Box(\pi(\bm{T}_{cw} \bm{T}_{wo} \bm{X}_o)) \\
Box(\bm{X}) &= [\min_x \bm{X}, \max_x \bm{X}, \min_y \bm{X}, \max_y \bm{X}] \,.
\end{align}
The transformation $\bm{T}_{cw} \bm{T}_{wo}$ brings sampled surface points of the super-quadric in canonical coordinates, $\bm{X}_o = S(\bm{\alpha}, \epsilon_1, \epsilon_2)$, into camera coordinates before projecting them into the image using the perspective projection function $\pi$. We obtain $\bm{X}_o $ using the equal-distance sampling techniques of super-quadrics~\cite{pilu1995equal}. 
We model the prior object scale distribution of each object category as $P(\bm{\theta})=\mathcal{N}(\bm{\alpha} | \bm{\mu}_0, \bm{\Sigma}_0)$ using a multi-variate Gaussian distribution.
Ideally this prior would capture the uncertainty of the averaged detector-predicted 3D BB $\bm{\mu}_0$ for proper Bayesian MAP estimation. While there are ways to train CNNs to produce uncertainty estimates~\cite{kendall2017uncertainties}, we found that we can simply use the variance $\bm{\Sigma}_0$ of the scale distribution of each object category in Scan2CAD~\cite{avetisyan2019scan2cad} as a proxy. 
Intuitively, since the detector is trained on this distribution, $\bm{\Sigma}_0$ is a conservative upper-bound on the variance---a well trained 3D BB detector should do significantly better.



\begin{table*}[th]
    \centering
    \scriptsize
    \setlength{\tabcolsep}{3pt}
    \begin{tabular}{l|c c c c c c c c | c}
        \makecell{Prec./Rec./F1 \\ IoU$>0.25$} & bathtub  & bookshelf & cabinet & chair & display & sofa & table & trashbin & avg.\\
        \hline
        Vid2CAD~\cite{maninis20vid2cad} & 45.5/30.0/36.1 & 18.0/12.7/14.9 & 46.3/\textbf{34.6}/\textbf{39.6} & \textbf{70.1}/78.6/\textbf{74.1} & \textbf{44.1}/42.8/\textbf{43.5} & 40.8/45.1/42.8 & 46.6/50/2/48.3 & 60.2/37.9/46.5 & 56.1/54.5/55.2\\
        MOLTR~\cite{li2020mo} & \textbf{67.5}/\textbf{41.6}/\textbf{51.5} & 42.8/21.3/28.4 & 62.7/22.8/33.5 & 58.7/77.4/68.6 & 17.7/34.5/23.4 & 69.4/52.2/59.5 & 63.5/57.4/60.3 & 49.0/42.6/45.6 & 54.2/55.8/55.0\\
        \methodname (ours) & 58.6/34.2/43.2 & \textbf{52.0}/\textbf{25.1}/\textbf{33.7} & \textbf{63.0}/26.4/37.2 & 68.3/\textbf{78.7}/73.1 & 37.5/37.5/37.5 & \textbf{75.9}/\textbf{53.1}/\textbf{62.5} & \textbf{65.5}/\textbf{58.9}/\textbf{62.0} & \textbf{67.8}/\textbf{60.8}/\textbf{64.1} & \textbf{64.7}/\textbf{58.6}/\textbf{61.5}\\
        \hline
        \makecell{ IoU$>0.5$} \\
        \hline
        Vid2CAD~\cite{maninis20vid2cad} & 2.5/1.6/2.0 & 0.0/0.0/0.0 & 7.7/5.7/6.6 & 29.2/32.8/30.9 & 0.0/0.0/0.0 & 0.8/0.8/0.8 & 6.7/7.2/6.9 & 23.2/14.6/17.9 & 16.8/16.3/16.5\\
        MOLTR~\cite{li2020mo} & 10.3/6.6/8.1 & 8.6/4.7/6.1 & 19.6/8.1/11.5 & 20.0/28.4/23.5 & 1.8/4.1/2.5 & 20.0/15.9/17.7 & 12.1/11.7/11.9 & 13.0/12.9/12.9 & 15.2/17.1/16.0\\
        \methodname (ours) & \textbf{14.3}/\textbf{8.3}/\textbf{10.5} & \textbf{11.5}/\textbf{5.7}/\textbf{7.6} & \textbf{25.9}/\textbf{10.9}/\textbf{15.3} & \textbf{39.0}/\textbf{44.8}/\textbf{41.7} & \textbf{7.7}/\textbf{7.7}/\textbf{7.7} & \textbf{39.2}/\textbf{27.4}/\textbf{32.3} & \textbf{26.0}/\textbf{23.3}/\textbf{24.6} & \textbf{31.6}/\textbf{28.0}/\textbf{29.5} & \textbf{31.2}/\textbf{28.3}/\textbf{29.7}\\
    \end{tabular}
    \caption{Quantitative ScanNet evaluation. \methodname outperforms MOLTR~\cite{li2020mo} and Vid2CAD~\cite{maninis20vid2cad} in four classes at IoU$>0.25$ and all classes at IoU$>0.5$ respectively.}
    \label{tab:vid2cad}
\end{table*}

\subsection{Implementation}\label{subsec:details}
\noindent\textbf{Detector training.}
Our detector is built upon DETR~\cite{carion2020detr}, a state-of-the-art 2D object detector that predicts objects as a set without post-processing.
We add three additional heads to DETR, each of which comprises three 512-dimension fully-connected layers, for object depth, 3D dimensions, 3D BB orientation respectively.
We fine-tune our detector from the pre-trained network weights on MSCOCO dataset~\cite{lin2014microsoft} for 10 epochs using ScanNet images and Scan2CAD annotations.
Although we use DETR in this work, other detectors such as MaskRCNN~\cite{he2017mask} can also be adopted.

\noindent\textbf{Graph neural network details.} 
A 3-layer MLP encoder is used to map the input to a 256D feature vector. 
The detection fusion block contains four self-attention layers producing 256D fused features.
The matching network for the fused features and frame detections is similar to SuperGlue~\cite{sarlin2020superglue} except we use six alternating cross- and self-attention layers.

\noindent\textbf{Optimization details.}
All parameters of the super-quadrics except $\epsilon_{1,2}$ which are initialized to $1$, are initialized using the average of associated single-view 3D prediction.
We sample
$1000$ points on the super-quadric surface for the optimization and found an assumed image variance of the 2D BB detector of $\sigma^2 = 20$ worked well.
We use the Adam optimizer in Pytorch to optimize the logarithm of the posterior in Eq.~\eqref{eq:map} for 20 iterations for every 50 associated 2D observations followed by a final optimization for 200 iterations at the end of the sequence.

\section{Experiments}

We evaluate the performance of our object-based mapping using the precision and recall metrics on ScanNet~\cite{dai2017scannet} and Scan2CAD~\cite{avetisyan2019scan2cad}. Because the original annotations do not provide amodal 3D BBs, following prior art~\cite{li2020mo, maninis20vid2cad}, we use the amodal 3D BB annotations from Scan2CAD as ground-truth. The precision is defined as the percentage of estimated super-quadrics being close enough to an annotated ground-truth 3D BB. The recall is the percentage of ground-truth 3D BBs that are covered by an estimated super-quadric. 
Specifically, a super-quadric is considered a true positive if the Intersection-over-Union (IoU) between its minimum enclosing 3D oriented 3D BB and a ground-truth BB in the same object class is above a pre-defined threshold. We use $0.25$ and $0.5$ in our experiments. 
A ground-truth BB can only be matched once to penalize repeated objects.
Note that we do not use mean Average Precision (mAP) which is typically used for the object detection because the proposed system outputs an unordered set of 3D bounding volumes without confidence scores.



\subsection{Comparing with RGB-only Methods}
\begin{figure*}[h]
    \centering
    \newcolumntype{Y}{>{\centering\arraybackslash}X}
    \begin{tabularx}{\linewidth}{@{}Y@{\,}Y@{\,}Y@{\,}Y@{}}
        \includegraphics[width=\linewidth]{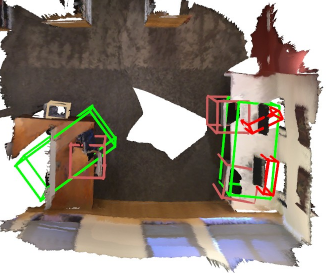}&
        \includegraphics[width=\linewidth]{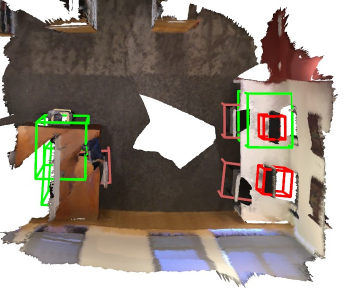}&
        \includegraphics[width=\linewidth]{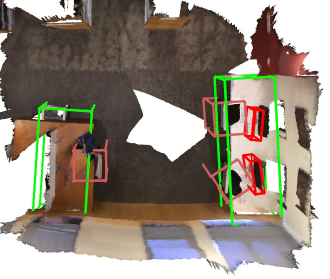}&
        \includegraphics[width=\linewidth]{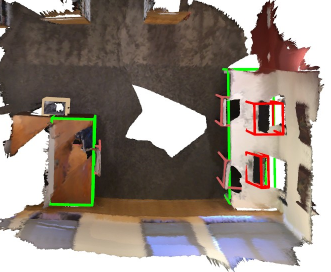}\\
        \includegraphics[width=0.8\linewidth]{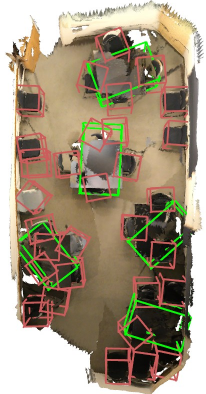}&
        \includegraphics[width=0.8\linewidth]{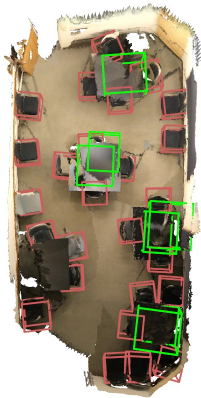}&
        \includegraphics[width=0.8\linewidth]{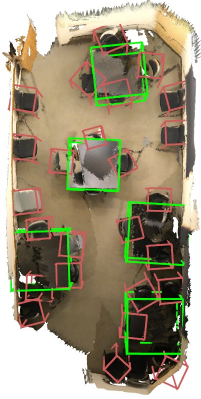}&
        \includegraphics[width=0.8\linewidth]{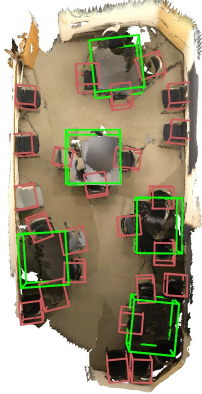}\\
        \includegraphics[width=0.9\linewidth]{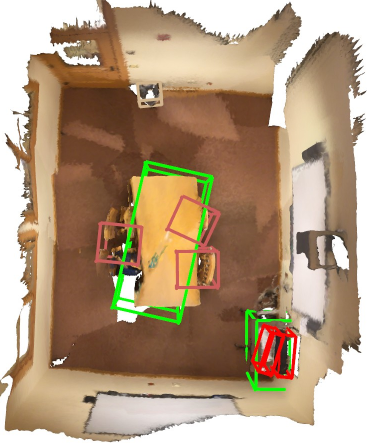}&
        \includegraphics[width=0.9\linewidth]{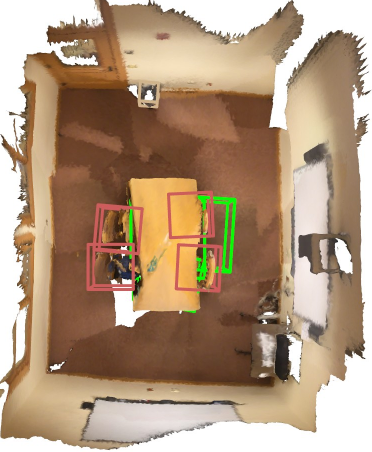}&
        \includegraphics[width=0.9\linewidth]{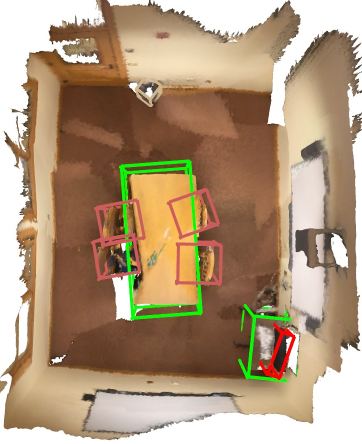}&
        \includegraphics[width=0.9\linewidth]{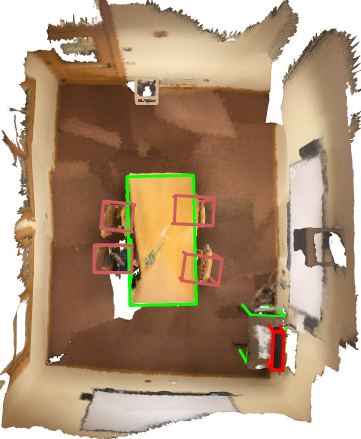}\\
         MOLTR~\cite{li2020mo} &
         Vid2CAD~\cite{maninis20vid2cad} &
         \methodname (Ours) &
         GT
    \end{tabularx}
    \caption{Qualitative comparison on ScanNet sequences. The colors of 3D BBs denote different categories. Both Vid2CAD and MOLTR suffer from replicated objects due to data association failure. 3D bounding boxes from our method are closer to ground-truth boxes thanks to our robust multi-view optimization.}
\label{fig:scan2cad_result}
\end{figure*}

We compare \methodname against two previous posed RGB videos methods, Vid2CAD~\cite{maninis20vid2cad} and MOLTR~\cite{li2020mo}, on ScanNet. 
These methods use are handcrafted data association (3D GIoU in MOLTR and Vid2CAD uses a combination of 2D IoU and visual appearance). MOLTR does not use multi-view geometry but fuses monocular 3D predictions via a filter in 3D, and the multi-view optimization in Vid2CAD lacks the scale prior in ours to alleviate the effect of inaccurate 2D observations. In contrast, we use attention-based GNN for association, followed by multi-view optimization.
Tab.~\ref{tab:vid2cad} shows precision, recall, and F1 score comparison per class at IoU thresholds of $0.25$ and $0.5$. Overall, ODAM outperforms Vid2CAD and MOLTR by about $6\%$ at $\text{IoU}>0.25$, and about $14\%$ at $\text{IoU} > 0.5$. 
As shown in Fig.~\ref{fig:scan2cad_result}, we can see duplicated objects in MOLTR and Vid2CAD due to failure in data association. 
Notably, our multi-view optimization estimates accurate oriented bounding volumes of large objects (\eg tables), wheres MOLTR and Vid2CAD often produce misaligned results.



\begin{table}[t]
    \centering
    \scriptsize
    \begin{tabular}{l | c c c | c}
        Methods & \multicolumn{3}{c|}{Components} & matching \\
        & GNN & monocular 3D & F2M association & accuracy\\
        \hline
        \multirow{3}{*}{Baselines} & \ding{51} & \ding{51} & \ding{55} & 0.86 \\
        & \ding{51} & \ding{55} & \ding{51} & 0.84 \\
        & \ding{55} & \ding{51} & \ding{51} & 0.85 \\
        \hline
        \methodname (ours) & \ding{51} & \ding{51} & \ding{51} & \textbf{0.88} \\
    \end{tabular}
    \caption{Ablation study on the learned data-association component of \methodname.
    The full model using GNN, monocular 3D detection, and frame-to-model (F2M) association achieves the best result.}
    \label{tab:data_association}
    \vspace{-0.25cm}
\end{table}


\subsection{Ablation study}\label{subsec:exp_ablation}
We validate the design choices in all key parts of \methodname using three ablation studies.

\begin{figure*}
    \centering
    \includegraphics[width=0.9\linewidth]{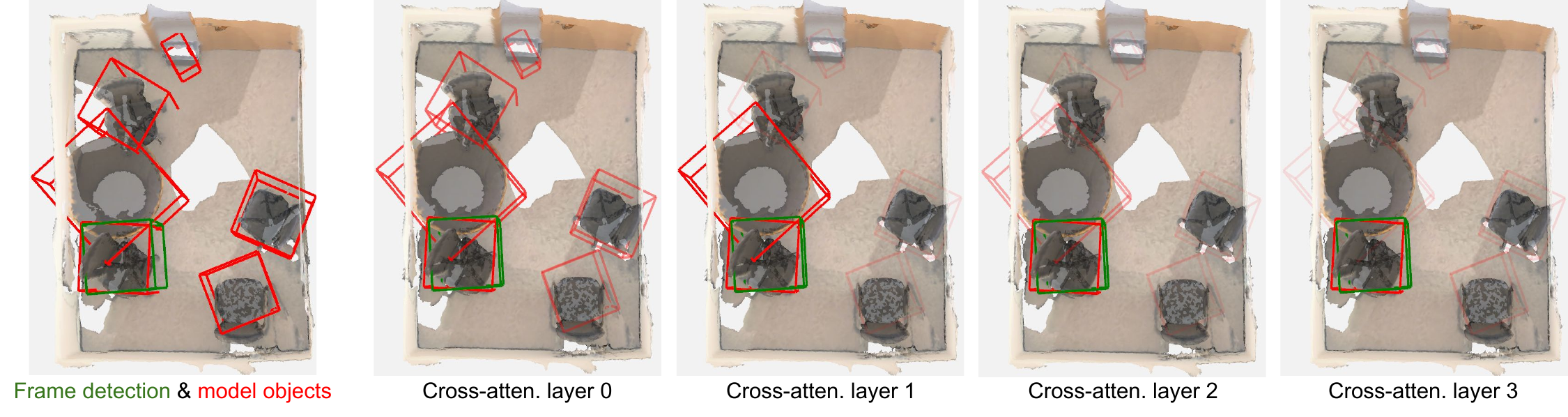}
    \caption{Visualization of GNN attention. The cross-attention scores of the 3D detection from the current frame detection (green) and model objects (red) are shown across various layers. Higher attention scores correspond to more opaque red BBs. The spread of the cross-attention shrinks and focuses on the correct assignment in deeper layers of the GNN.}
    \label{fig:attention_demo}
\end{figure*}

\noindent\textbf{Data association.}
The key aspects we consider in this ablation study are: (1) GNN vs. handcrafted pairwise cost, (2) the effect of single-view 3D attribute estimation vs. 2D only attribute in data association, (3) frame-to-model association vs. frame-to-frame association.
When the GNN is not used, we use 3D BB IoU as the matching cost following Li \etal~\cite{li2020mo}. 
To validate the importance of the detection fusion block in the GNN (\ie frame-to-model association), we compare it against a baseline GNN which only takes as input the latest observation of existing object instances, which can be considered as a frame-to-frame association.
Besides the final 3D mapping result, we also use the matching accuracy as a direct measurement for the data association algorithms. 
Tab.~\ref{tab:data_association} shows all three key components contribute to the performance gain. 
Fig.~\ref{fig:attention_demo} visualizes how the attention scores change across different layers in the GNN.

\noindent\textbf{Shape representation.}
Tab.~\ref{tab:ablation} shows that optimizing with the super-quadric representation performs better than cuboid and ellipsoid by $2.5\%$ and $9\%$, respectively. 
Cuboid outperforms ellipsoid because a considerable amount of objects are cuboid-like in the evaluated object classes.
Further qualitative comparisons can be found in the supplement. 

\noindent\textbf{Optimization.}
Tab.~\ref{tab:ablation} shows the effect of the back-end multi-view optimization and the scale prior terms in the objective function.
The ``no optimization'' results are obtained by taking the average of associated monocular 3D predictions without any multi-view optimization and have the worst in the group. 
This indicates that single-view 3D detector alone is not sufficient for object-based mapping.
Using only 2D bounding box observations for the multi-view optimization is also suboptimal, giving a minor $1.8\%$ deterioration. 
Our full approach (using 2D bbox and prior jointly) outperforms the ``no optimization'' baseline by $5.8\%$.
To better demonstrate how the errors in 2D BBs affect the optimization, we show how the performance gap between optimization w/ and wo/ the prior term changes as the errors in the 2D BBs increase in the supplement.

\subsection{Comparing with RGB-D methods}

\begin{table}[t]
    \centering
    \footnotesize
    \begin{tabular}{ c | c || c}
    \multicolumn{2}{c||}{Switched component} &  Result (Prec./Rec./F1) \\
    \hline\hline
    \multirow{2}{*}{Shape representation} &  ellipsoid & 21.9/19.6/20.7 \\
    & 3D cuboid & 28.5/26.1/27.2 \\
    \hline
    \multirow{2}{*}{Optimization} & no optimization & 25.2/22.8/23.9 \\
    & wo/ scale prior & 22.9/21.3/22.1 \\
    \hline
    \hline
    \multicolumn{2}{c||}{\methodname (ours)} & \textbf{31.2}/\textbf{28.3}/\textbf{29.7}
    \end{tabular}
    \caption{Ablation study on different shape representations and the multi-view optimization. The combination of super-quadric representation and scale-prior in the multi-view optimization leads to the best performance.}
    \label{tab:ablation}
    \vspace{-0.5cm}
\end{table}

\begin{table*}[t]
    \centering
    \footnotesize
    \begin{tabular}{l|c c c c c c c c c c c c c c c c c}
        F1 & \rotatebox{90}{cabinet} & \rotatebox{90}{bed} & \rotatebox{90}{chair} & \rotatebox{90}{sofa} & \rotatebox{90}{table} & \rotatebox{90}{door} & \rotatebox{90}{window} & \rotatebox{90}{boohshelf} & \rotatebox{90}{picture} & \rotatebox{90}{counter} & \rotatebox{90}{desk} & \rotatebox{90}{curtain} & \rotatebox{90}{fridge} & \rotatebox{90}{shower} & \rotatebox{90}{toilet} & \rotatebox{90}{bath} & \rotatebox{90}{others} \\
        \hline
        VoteNet~\cite{qi2020imvotenet} & \textbf{40.9} & \textbf{88.1} & \textbf{85.2 }& \textbf{79.8} & 60.0 & \textbf{53.0 }& \textbf{40.7} & \textbf{50.0} & \textbf{12.5} & \textbf{52.3} & 62.2 & \textbf{50.0} & 47.4 & \textbf{50.5} & 90.7 & 53.9 & \textbf{91.5} \\
        \methodname (ours) & 22.1 & 87.7 & 74.9 & 61.8 & \textbf{65.6} & 12.5 & 12.8 & 40.5 & 7.4 & 9.3 & \textbf{65.0} & 13.1 & \textbf{48.7} & 41.2 & \textbf{93.1} & \textbf{64.8} & 83.3 \\
    \end{tabular}
    \caption{Comparison to VoteNet~\cite{qi2020imvotenet}. VoteNet relies on colored 3D point cloud which greatly simplifies 3D object localization. \methodname performs similarly in most categories but struggles with thin objects such as door, window and curtain.}
    \label{tab:votenet}
    \vspace{-0.25cm}
\end{table*}

This comparison is to identify the current gap between RGB and RGB-D methods. 
We compare to VoteNet~\cite{qi2020imvotenet}, a state-of-the-art 3D object detection network using colored point clouds.
Compared to RGB-only, the additional depth information, which is fused into a point cloud before 3D object detection, significantly simplifies the task. The 3D structure is explicitly represented and becomes an input to the 3D object-detection system and does not have to be inferred by the system.
Yet the RGB-only methods are valuable because depth sensors consume additional power and most consumer-grade devices have limited range.

We train our detector and the GNN using the original ScanNet annotations to be consistent with VoteNet. We select the score threshold in VoteNet that leads to the best F1 score. As shown in Tab.~\ref{tab:votenet}, we achieve comparable or even superior performance to VoteNet in some object classes (\eg bed, table, desk, fridge, toilet, and bath). 
This is because these objects are normally arranged distantly to other object instances in the same class, making the data association easier.
On the other hand, our method struggles with thin objects, such as door, window, picture, and curtain, because a small localization error results in a significant drop in 3D IoU leading to worse F1 score.

\subsection{Run-time analysis}
All experiments are run on a Nvidia GeForce GTX 1070 GPU. 
The monocular detector can run at about 10~fps. 
Although the inference time of the GNN grows linearly with the number of objects in the map, the GNN runs at 15 fps on average in all ScanNet validation sequences.
Overall, the front-end of \methodname can achieve around 6~fps.
A naive back-end optimization using the Pytorch Adam optimizer takes $0.2$ seconds for 20 iterations. 
This back-end optimization is not time critical and can be run in a parallel thread. 
It could also be accelerated significantly using second order methods such as implemented in GTSAM~\cite{dellaert2012factor}.



\section{Conclusion}
We presented \methodname, a system to localize and infer 3D oriented bounding volumes of objects given posed RGB-only videos.
Key to \methodname is (1) an attention-based GNN for robust detection-to-map data association, and (2) a super-quadric-based multi-view optimization for accurate object bounding volume estimation from the associated 2D BB and class observations.
\methodname is the best performing RGB-only method for object-based mapping.
The fact that the proposed RGB-only methods can close the accuracy gap to RGB-D methods in a subset of object categories is encouraging and points to a future where depth cameras are unnecessary for 3D scene understanding. 

\noindent \textbf{Acknowledgment} KL and IR gratefully acknowledge the support of the ARC through the Centre of Excellence for Robotic Vision CE140100016.
\newpage

{\small
\bibliographystyle{ieee_fullname}
\bibliography{reference}
}
\end{document}


\title{ODAM: Object Detection, Association, and Mapping using Posed RGB Video}

\makeatletter
\renewcommand\AB@affilsepx{, \protect\Affilfont}
\makeatother

\author[1, 2]{Kejie Li}
\author[2]{Daniel DeTone}
\author[2]{Steven Chen}
\author[2]{Minh Vo}
\author[1]{Ian Reid}
\author[3]{Hamid Rezatofighi}
\author[2]{Chris Sweeney}
\author[2]{Julian Straub}
\author[2]{Richard Newcombe}

\affil[1]{The University of Adelaide}
\affil[2]{Facebook Reality Labs}
\affil[3]{Monash University}

\maketitle

\noindent\textbf{GNN details.}
The object feature descriptor is mapped to a 256-dimensional embedding using the 3-layer MLP encoder, the output dimensions of which are $64$, $256$, and $256$, before being processed by the GNN. 
After the encoding, every node in the graph is described by a 256-dimensional feature vector. 
An attention layer of the GNN takes as input node features of the last layer and outputs the updated node features by aggregating information from other nodes. 
Specifically, the message passing among nodes is achieved by self-attention or cross-attention depending on the connection type among the nodes (lines 281-285 in the main text).

The update of each node feature in an attention layer is proceeded as follows: (1) For each node in the graph, we employ a 4-head attention mechanism to aggregate information from other nodes; 
(2) The aggregated feature is then concatenated with the node feature;
(3) The concatenated feature is passed to a 3-layer MLP (with dimensions of $512$, $512$, $256$) to update the node feature.
In the second part of the GNN for frame-to-model association, we we use the optimal matching layer~\cite{sarlin2020superglue} to obtain the assignment matrix. 
We train the GNN in a supervised fashion using the ground-truth assignments by minimizing the negative log likelihood the correct assignment:
\begin{align}
&\begin{aligned}
    L = &-\sum_{(i, j)\in \mathcal{S}} log \hat{\bm{M}}_{i,j} - \sum_{i \in \mathcal{S}_0}log\hat{\bm{M}}_{i, n+1} \\
    &-\sum_{j \in \mathcal{S}_1}log\hat{\bm{M}}_{m+1, j}\,,
&\end{aligned}
\end{align}
where $\hat{\bm{M}} \in \mathbb{R}^{m+1, n+1}$ (the extra one dimension for the dustbin~\cite{sarlin2020superglue}) is the assignment prediction, $\mathcal{S}$ is the ground-truth matching pairs, and $\mathcal{S}_0$ and $\mathcal{S}_1$ are the objects or detections that are not matchable due to occlusion or out-of-frame, which should be assigned to the dustbin.

\begin{figure}
    \centering
    \includegraphics[width=\linewidth]{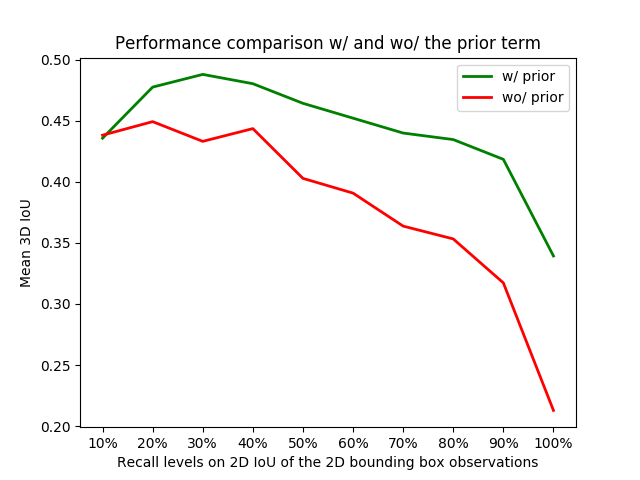}
    \caption{Comparison of 3D IoU between optimization w/ the prior term (in green) and wo/ the prior term (in red) against the 2D observation errors. Optimization with the prior term is less affected by the errors in the 2D observations.}
    \label{fig:prior_comparison}
\end{figure}

\noindent\textbf{The effect of the prior term.}
Besides reporting the overall performance gain due to the prior term in the optimization (see Table 3 in the main text), we demonstrate the performance difference between optimization w/ prior and wo/ prior in different levels of 2D observation errors in this section.
We rank the predicted 3D objects using the mean 2D IoU between the associated 2D bounding boxes and the ground-truth bounding boxes in descending order, and plot the mean 3D detection performance measured by 3D IoU at different levels of 2D observation errors. 
Fig.~\ref{fig:prior_comparison} shows that as the error in 2D observations increase, the performance of optimization wo/ prior drops significantly whereas the optimization w/ prior is less affected, which further validates that the prior term can increase robustness of the multi-view optimization to error in 2D observations.

\begin{figure*}
    \centering
    \includegraphics[width=\linewidth]{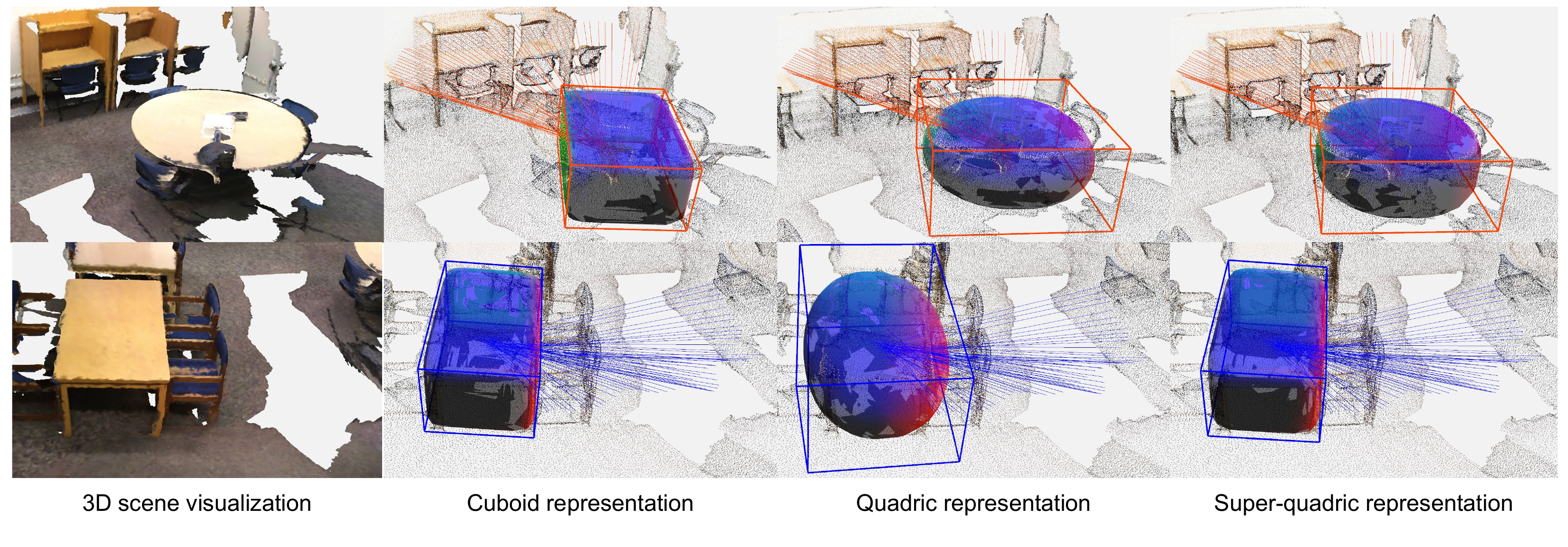}
    \caption{Visualization of cuboid, quadric, and super-quadric representation. The super-quadric representation can adapt to different object shapes while cuboid or quadric can only fit box-like and round objects well respectively.}
    \label{fig:representation_limit}
\end{figure*}

\noindent\textbf{Representation comparison.}
Fig.~\ref{fig:representation_limit} shows some examples demonstrating the limitation of cuboid or quadic representation. 
Although cuboid seems to be more favorable than ellipsoid as reported in Table 3 in the main text, one should note that most object classes in the Scan2CAD annotations for evaluation are box-like furniture. Ellipsoid would be advantageous for round or cylinder objects, such as cups, fruits, and balls.
Super-quadric, as a unified representation for shapes including but not limited to cuboids, cylinders, and ellipsoids, is a more flexible representation for generic object shapes, as shown in Fig.~\ref{fig:representation_limit} and Table 3 in the main text.

\begin{figure*}
    \centering
    \includegraphics[width=\linewidth]{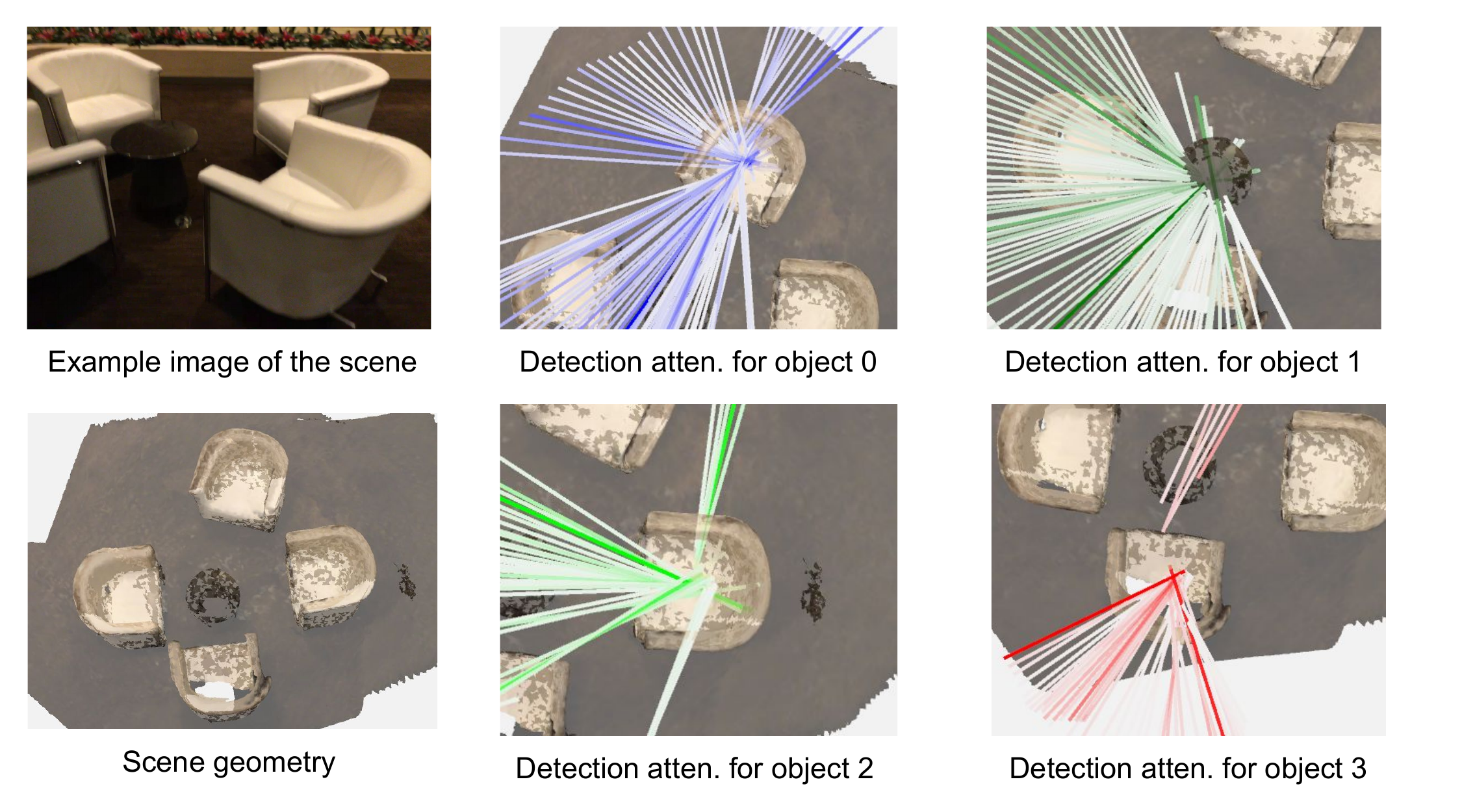}
    \caption{Visualization of object fusion attention. Each line represents a previously associated detection of an object. The attention score of a detection used for fusion is represented by the intensity of the color. The network learns to attend to detections from various viewpoints.}
    \label{fig:detection_fusion}
\end{figure*}

\noindent\textbf{Self-attention visualization.}
Fig.~\ref{fig:detection_fusion} visualizes the attention weights of the object fusion block in the GNN (described at line 304 in the main text).
It is interesting that the network focuses on a subset of observations with a large viewpoint difference.

\noindent\textbf{More qualitative results and failure case analysis.}
More qualitative results including failure cases are shown in the supplementary video.

{\small
\bibliographystyle{ieee_fullname}
\bibliography{reference}
}